\documentclass[journal]{journal}

\usepackage{helvet}
\usepackage{courier}
\usepackage[hyphens]{url}
\usepackage{graphicx}
\usepackage{caption}
\usepackage{algorithm}
\usepackage{algpseudocode}
\usepackage{amsmath}

\usepackage{newfloat}
\usepackage{listings}
\usepackage{comment}
\usepackage{subcaption}
\usepackage{xcolor}
\usepackage{dashrule}
\usepackage{tabularx}
\usepackage{booktabs}   
\usepackage{amsfonts}   
\usepackage{ragged2e}
\usepackage{cite}

\definecolor{custompurple}{HTML}{800080}  
\definecolor{customblue}{HTML}{1F77B4}
\definecolor{customgreen}{HTML}{66C2A5}
\definecolor{custombrown}{HTML}{654321}  
\definecolor{dodgerblue}{HTML}{1E90FF}  
\definecolor{customgrey}{HTML}{8DA0CB}
\definecolor{customorange}{HTML}{FC8D62}
\definecolor{CCLF}{HTML}{A6D854}
\usepackage{hyperref}

\begin{document}
\title{Integrating Novelty and Surprise for Experience Prioritization and Exploration in Image-Based Reinforcement Learning}

\author{Hoda Yamani, Henry Williams, Bruce A. MacDonald%
\thanks{
Robot Learning Team and CARES Robotics Lab,  
Department of Electrical, Computer, and Software Engineering,  
University of Auckland, Auckland, New Zealand.  
Corresponding author: hyam650@aucklanduni.ac.nz.
}%
}
\maketitle

\begin{abstract}
Sample efficiency is a central challenge in reinforcement learning (RL), particularly in image-based domains where agents must learn from high-dimensional visual inputs. Traditional sampling often relies on random or suboptimal experience selection, leading to redundant updates and slow learning. Improving efficiency requires mechanisms that prioritize informative experiences while also encouraging effective exploration. Prioritized Experience Replay (PER) addresses part of this challenge by reusing high-value transitions, while intrinsic rewards promote the exploration of novel or uncertain states. However, their integration has not been extensively studied. This paper introduces Novelty and Surprise Prioritized Experience Replay (NSPER), which uses novelty to capture underrepresented states and surprise to expose gaps in the agent’s understanding of the environment. We further extend this with NSPER+R, integrating these signals as intrinsic rewards to jointly improve replay quality and exploration. Experiments on DeepMind Control Suite tasks show that NSPER and NSPER+R improve training efficiency and convergence speed compared to existing methods in image-based RL. 

\end{abstract}

\begin{figure*}[!t]
    
        \centering
    
        \includegraphics[width=0.8\textwidth]{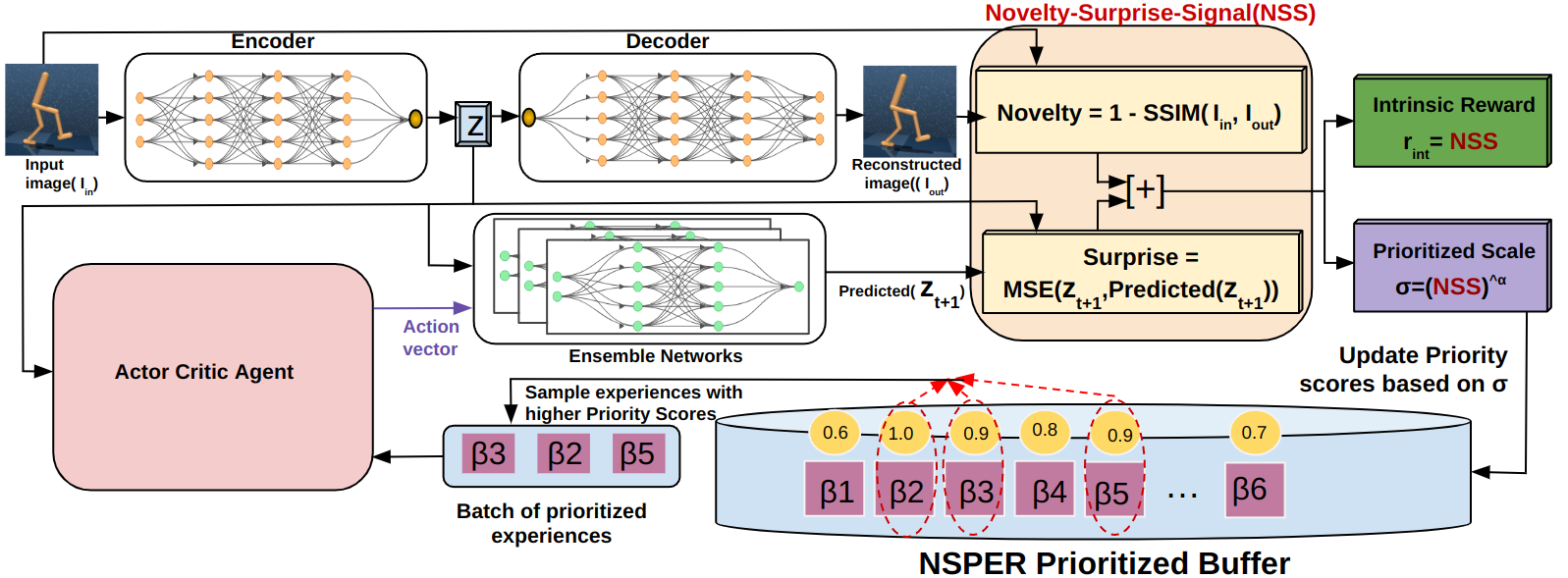}
    
       \caption[A high-level representation of NSPER]{Schematic overview of Novelty and Surprise Prioritized Experience Replay (NSPER).}

        \label{fig:NSPER}
    
    \end{figure*}
\section{INTRODUCTION}

    Reinforcement learning (RL) has shown great promise in image-based applications, allowing agents to learn directly from visual input \cite{10611330,10377012}. This paradigm has been widely applied in domains such as gaming \cite{mnih2016asynchronous}, robotics \cite{ferraro2022computational}, and autonomous driving \cite{scorsoglio2020image}, where vision serves as the primary source of environmental feedback. However, high-dimensional image data introduces redundancy, high memory demands, and substantial computational overhead, making policy learning more complex and resource-intensive \cite{chen2021improving}. Since RL inherently relies on large volumes of interaction data, improving sample efficiency is critical \cite{dorner2021measuring}. A promising direction is to prioritize informative experiences during training and to guide exploration toward those experiences that provide the most learning value. This prioritization can be guided by intrinsic learning signals that highlight which transitions are the most useful for learning \cite{barto2013intrinsic}.
    
     Inspired by human cognition, where intrinsic signals foster curiosity \cite{baldassarre2011intrinsic,deci2013intrinsic} and guide memory systems to prioritize novel or salient experiences \cite{shohamy2010dopamine,braun2018retroactive}, RL agents can adopt similar mechanisms to enhance exploration and accelerate learning. Such signals provide a foundation for both directing exploratory behavior and selecting the most informative experiences during training \cite{pathak2017curiosity,houthooft2016vime}, thereby improving sample efficiency in two complementary ways. First, they can be formulated as intrinsic rewards, encouraging agents to acquire diverse and informative experiences that accelerate policy refinement \cite{oudeyer2007intrinsic}. Second, they can be used as prioritization scores within experience replay, enabling agents to focus on transitions with the highest learning potential.
     
     Among intrinsic signals, novelty and surprise have proven especially effective as intrinsic rewards \cite{bellemare2016unifying}. Novelty reflects the unfamiliarity of a state, driving exploration toward underrepresented or unexplored regions \cite{jaegle2019visual,kubovvcik2023signal}. Surprise, on the other hand,  arises when outcomes deviate substantially from predictions, exposing gaps in the agent’s environmental model and its limited ability to generalize from past experiences \cite{achiam2017surprise}. Combining these two signals enables agents to identify regions that require further learning and adjust their exploration strategies, resulting in more efficient and targeted exploration \cite{le2024beyond,valencia2024image}.

     Prioritized Experience Replay (PER) is a well-established approach in RL that exploits such intrinsic signals within the replay buffer to improve sample efficiency \cite{schaul2015prioritized}. In its original form, which we refer to as TD-PER, transitions are prioritized based on the temporal difference (TD) error, which measures the discrepancy between predicted and actual outcomes following an agent’s action. While effective in discrete action spaces, TD-PER struggles in continuous control environments \cite{pan2022understanding,oh2021model}. To overcome these limitations, several extensions have been proposed, including augmenting TD error with auxiliary signals or using biologically inspired mechanisms for prioritization \cite{saglam2022actor, oh2021model, yamani2025reward, sun2022cclf}. Despite these advances, integrating intrinsic rewards with PER in image-based RL remains limited. For instance, the Contrastive Curiosity Learning Framework (CCLF) \cite{sun2022cclf} uses curiosity to prioritize surprising augmented inputs. However, the broader use of intrinsic rewards as scoring signals for experience prioritization remains largely underexplored.

    To address this gap, we propose NSPER, a novel prioritization method that uses two distinct intrinsic signals, novelty and surprise, to guide experience selection within PER. By combining these signals to guide sampling, NSPER identifies diverse and informative transitions with high learning potential, improving the quality of replayed experiences. We further introduce NSPER+R, an extension of NSPER that integrates novelty and surprise not only for prioritization but also as intrinsic rewards. This dual mechanism improves experience selection and exploratory behavior, leading to faster convergence and more generalizable policies. To evaluate our approach, we integrate it into PixelTD3, an image-based variant of the TD3 algorithm \cite{fujimoto2018addressing}, and assess its performance on challenging tasks from the DeepMind Control Suite \cite{tassa2018deepmind}, demonstrating promising improvements in complex, high-dimensional visual environments. The model architecture is illustrated in Fig.~\ref{fig:NSPER}. The implementation is publicly available at \url{https://github.com/UoA-CARES/NSPER}.
    
    The main contributions of this paper are summarized as follows:
    
    \begin{enumerate}
    \item We introduce \textbf{NSPER}, a novel approach that uses intrinsic rewards as prioritization signals within PER to improve sample efficiency in image-based RL.  

    \item We extend this idea in \textbf{NSPER+R}, which uses novelty and surprise both for prioritization and as intrinsic rewards, further enhancing exploration and policy learning.

    \item We provide a comprehensive evaluation of different prioritization strategies and their integration with intrinsic rewards, analyzing their impact on training efficiency and learning performance.
    
    \item We conduct an ablation study to isolate the individual contributions of novelty and surprise, as well as their combined effect when used with intrinsic rewards.
    
    \end{enumerate}



\section{RELATED WORKS}
Experience prioritization and intrinsic motivation are key strategies for improving sample efficiency and exploration in RL. Prior work has explored various mechanisms for identifying informative experiences and guiding agents toward novel or surprising transitions, providing a foundation for methods that combine these ideas to enhance learning in complex environments.

\subsection{\textbf{Experience Prioritization Methods}}
   PER methods enhance sample efficiency by assigning higher sampling probability to transitions with greater learning potential \cite{zha2019experience}. TD-PER \cite{schaul2015prioritized}, a common variant, ranks transitions based on TD error, under the assumption that transitions with larger TD errors are less familiar and therefore more informative. However, in continuous high-dimensional spaces, small perturbations in actions or observations can lead to large fluctuations in value estimates, even for well-learned transitions, making TD error a noisy and unstable prioritization signal \cite{pan2022understanding}. This sensitivity makes TD error a noisy and unreliable prioritization signal, motivating alternative strategies that offer more stable and informative learning signals in continuous control tasks \cite{carrasco2025uncertainty,liu2021regret}.
  
   To address these limitations, LA3P \cite{saglam2022actor} decouples learning signals by assigning low TD-error transitions to the actor to stabilize policy updates and high TD-error transitions to the critic to improve value estimation. Another method \cite{zha2019experience} prioritizes recent transitions with low TD error, thereby focusing on up-to-date experiences, but this risks catastrophic forgetting by discarding older yet valuable data. Model-Augmented PER (MaPER) \cite{oh2021model} incorporates auxiliary signals derived from learned environmental dynamics to complement the TD error, thereby improving the identification of important experiences. Despite these enhancements, all of these methods primarily rely on the TD error as the primary prioritization criterion.

    In contrast, Reward Prediction Error (RPE) prioritization \cite{yamani2025reward} introduces an alternative signal inspired by biological learning. By measuring the discrepancy between expected and actual rewards, RPE provides informative feedback on its prediction accuracy, allowing it to adjust its policies and enhance decision-making over time \cite{simmons2019reward}. Similarly, our method proposes a novel prioritization strategy that uses alternative intrinsic signals, fostering exploration and emphasizing unfamiliar transitions.
    
\subsection{\textbf{Intrinsic Motivation Strategies}}

     Intrinsic motivation has been widely explored in RL to encourage agents to explore and learn beyond extrinsic rewards ~\cite{oudeyer2007intrinsic}. This approach encourages agents to seek out novel states and behaviors within the environment \cite{aubret2019survey, sun2022cclf}. Several intrinsic motivation mechanisms have been proposed, including state novelty \cite{bellemare2016unifying}, state prediction errors \cite{bellemare2016unifying}, uncertainty about outcomes \cite{li2021mural}, and environment dynamics \cite{seo2021state}. Among these, novelty and surprise enable more focused and efficient learning by directly targeting meaningful opportunities, guiding exploration toward unfamiliar states and significant prediction errors, and outperforming methods that rely solely on uncertainty or environmental dynamics \cite{valencia2024image}.

     Novelty encourages visiting rarely encountered states, promoting a broader understanding of the environment \cite{jaegle2019visual}. Surprise is measured through prediction errors in environment dynamics, such as next-state errors and state transition divergences \cite{pathak2017curiosity,achiam2017surprise}. Combining novelty and surprise captures complementary aspects of agent-environment interaction, improving exploration and representation learning \cite{jaegle2019visual,kubovvcik2023signal} and allowing better balance between exploration and exploitation in complex tasks \cite{le2024beyond}.

     Despite these advances, integrating intrinsic motivation signals into memory prioritization mechanisms remains an ongoing challenge. Several recent studies have focused on combining curiosity-based metrics with experience replay to further refine prioritization strategies.

\subsection{Integrating Intrinsic Signals in Experience Replay}

    Recent studies have explored the incorporation of intrinsic motivation into memory prioritization. Curiosity-driven experience prioritization (CDP) \cite{zhao2019curiosity} prioritizes trajectories that lead to rare goal states, ensuring the agent focuses on underexplored areas. However, CDP is based on the replay of hindsight experiences (HER) \cite{andrychowicz2017hindsight}, which limits its applicability to goal-oriented tasks. Augmented Curiosity-Driven Experience Replay (ACDER) \cite{li2020acder} also integrates curiosity signals into HER but remains restricted to goal-conditioned exploration. 

    A more relevant approach, CCLF \cite{sun2022cclf}, introduces a self-supervised prioritization mechanism in image-based RL. CCLF selects surprising augmented inputs to enhance sample efficiency, but it focuses primarily on representation learning rather than on optimizing experience selection for policy improvement.

    Our method extends these insights by integrating novelty and surprise into experience prioritization, ensuring that selected transitions support both exploration and policy optimization. By combining intrinsic rewards with a structured prioritization mechanism, it overcomes the limitations of PER and leverages intrinsic motivation to enable more effective learning in challenging environments


\section{\textbf{METHODOLOGY}}
This section outlines the components underlying our approach, beginning with a review of PER, PixelTD3, and the computation of novelty and surprise signals. We then introduce our framework, NSPER, and its extension, NSPER+R.

\subsection{\textbf{PER}}

PER improves sample efficiency by replaying transitions with higher learning potential. Transitions in the buffer $B$ are stored as:
\begin{equation}\label{eq:1}
B_i = (s_i, a_i, r_i, s_{i+1}),
\end{equation}
where $i$ indexes the time step. The priority of each transition is defined as:
\begin{equation}\label{eq:2}
\sigma_i = |\delta_i| + \epsilon,
\end{equation}
In the standard variant, TD-PER, $\delta_i$ is the TD error, and $\epsilon > 0$ ensures non-zero priority. 

Transitions are sampled according to:

\begin{minipage}{0.48\linewidth}
\begin{equation}
p_i = \frac{\sigma_i^\alpha}{\sum_k \sigma_k^\alpha}
\label{eq:3}
\end{equation}
\end{minipage}%
\hfill
\begin{minipage}{0.48\linewidth}
\begin{equation}
w_i = \Big(\tfrac{1}{|B|\,p_i}\Big)^\beta
\label{eq:4}
\end{equation}
\end{minipage}

where $p_i$ favors transitions with higher priority $\sigma_i$, and $w_i$ corrects for the bias introduced by non-uniform sampling.  
Here, $\alpha$ controls prioritization sharpness, and $\beta$ compensates for sampling bias. While TD-PER defines $\sigma_i$ using TD error, PER is flexible and allows alternative prioritization signals by redefining $\sigma_i$.

\subsection{\textbf{PixelTD3}}

PixelTD3 is an image-based extension of TD3~\cite{fujimoto2018addressing}, inspired by SACAE~\cite{yarats2021improving} and extended in NaSATD3~\cite{valencia2024image}. PixelTD3 employs a convolutional autoencoder that maps raw observations $s_t$ into a compact latent representation $z$. This latent vector is shared between the actor and critic networks, ensuring that policy learning is grounded in a consistent feature space. The encoder--decoder is jointly trained with a reconstruction loss that preserves the structural content of the input images while maintaining task-relevant features.  

 Unlike SACAE, which trains separate encoders for actor and critic, PixelTD3 uses a single shared autoencoder that is updated end-to-end at every step, including both convolutional and linear layers.
 This provides a stable and compact representation for policy learning. We adopt this setup as the backbone of NSPER (see Fig.~\ref{fig:NSPER}). Complete architectural specifications of the encoder, decoder, actor, critics, and predictive ensemble are reported in Table~\ref{tab:arch-setup}.

\subsection{\textbf{Novelty And Surprise Computation}}

\textbf{Novelty} measures how different an observation is from what the agent has already encoded. It is estimated by reconstructing the input $s_t$ through the autoencoder and comparing it to the original observation. Let $\hat{s}_t = Dec(Enc(s_t))$ denote the reconstruction of $s_t$. The novelty score is then defined as the complement of the Structural Similarity Index (SSIM) between $s_t$ and $\hat{s}_t$:
\begin{equation}\label{eq:5}
\text{Novelty}(s_t) = 1 - \text{SSIM}(s_t, \hat{s}_t).
\end{equation}
A higher value indicates that the observation is less well-reconstructed, and therefore more novel to the agent. Fig.~\ref{fig:NSPER} illustrates how novelty is obtained from the autoencoder reconstruction error of the input image.

\textbf{Surprise} measures the deviation between the agent’s predicted and observed dynamics. It is computed in latent space using an ensemble of dynamics models $P_\rho$, which predict the next latent state $\hat{z}_{t+1}$ given the current latent representation $z_t$ and action $a_t$:
\begin{equation}\label{eq:6}
\hat{z}_{t+1} = P_\rho(z_t, a_t).
\end{equation}
Surprise is defined as the mean squared error between the predicted and observed latent states:
\begin{equation}\label{eq:7}
\text{Surprise}(s_t,a_t) = \text{MSE}(z_{t+1}, \hat{z}_{t+1}).
\end{equation}
As illustrated in Fig.~\ref{fig:NSPER}, this signal corresponds to the mismatch between the latent representation of the observed next state and the ensemble prediction. The ensemble consists of $M$ models trained with mean squared error loss, and averaging their predictions reduces variance and improves stability under non-stationary encoder updates.

\textbf{Novelty and Surprise Combination}: Combining novelty and surprise reward signals encourages more targeted exploration of informative regions within the environment. This integration forms a new intrinsic signal, referred to as the \textbf{Novelty-Surprise Signal (NSS)}. The \( \text{NSS}(s, a) \), derived from~\eqref{eq:5} and~\eqref{eq:7}, captures the joint contribution of novelty and surprise and is formulated as:

    \begin{equation}\label{eq:8}
    \text{NSS}(s_t, a_t) = \text{Novelty}(s_t) + \text{Surprise}(s_t, a_t)
    \end{equation}

\subsection{\textbf{NSPER: Novelty and Surprise Prioritization}}

We propose a prioritization strategy that leverages novelty and surprise as intrinsic signals to guide experience replay. Rather than relying on the TD error, as in standard PER, NSPER replaces $\delta_i$ in~\eqref{eq:2} with the Novelty–Surprise Signal $\text{NSS}(s,a)$ from~\eqref{eq:8}. The priority assigned to each transition is therefore defined as:
\begin{equation}\label{eq:9}
\sigma_t = \left|\text{NSS}(s_t, a_t)\right| + \epsilon,
\end{equation}
where $\epsilon > 0$ ensures all transitions remain eligible for replay.

These priorities are then used in the sampling equations~\eqref{eq:3} and~\eqref{eq:4} to compute transition probabilities and importance weights. By emphasizing transitions with higher novelty and surprise, NSPER favors diverse and informative experiences that better support critic training. The parameter $\alpha$ regulates the strength of this prioritization, ensuring that informative transitions are emphasized without over-concentrating sampling on a few extreme cases. This balance allows NSPER to prioritize informative experiences while maintaining diversity in the sampled transitions.

\subsection{\textbf{Enhancing Exploration with Intrinsic Rewards}}

Intrinsic rewards are agent-generated signals that promote exploration, especially when extrinsic rewards are sparse or delayed \cite{barto2013intrinsic}. In our approach, we define the intrinsic reward using novelty and surprise~\eqref{eq:8} as:

\begin{equation}\label{eq:10}
r_{int}(s_t, a_t) = NSS(s_t, a_t)
\end{equation}
As depicted in Fig.~\ref{fig:NSPER}, the NSS is used both to prioritize experiences during replay and to define the intrinsic reward signal.
The total reward at each step combines extrinsic and intrinsic components:

\begin{equation}\label{eq:11}
r_t = r_{ext} + r_{int}
\end{equation}

We refer to the full version of our framework as NSPER+R, which integrates intrinsic rewards with prioritized experience replay. This variant improves learning by prioritizing transitions that contribute to both task success and deeper environmental understanding. Algorithm \ref{alg:nss} summarises the full procedure for NSPER and its intrinsic-reward variant NSPER+R.
 \begin{algorithm}[H]
    \small
    \caption{NSPER and NSPER+R with NSS (Novelty–Surprise Signal)}
    \label{alg:nss}
    \begin{algorithmic}[1]
    \State \textbf{Inputs:} PER exponents $\alpha,\beta$, batch size $m$, small $\varepsilon>0$
    \State \textbf{Initialize:} actor $\Theta$, critic $\theta$, encoder--decoder $(\mathrm{Enc},\mathrm{Dec})$, replay buffer $B$, priorities $\{\sigma_i\}$
    
    \For{each environment step $t$}
      \State Observe state $s_t$; select action $a_t$ (with exploration noise); receive $r_{\text{ext}}$ and $s_{t+1}$
      \State \textbf{Novelty:} $\displaystyle \mathrm{Novelty}(s_t) \gets 1 - \mathrm{SSIM}\!\big(s_t,\, \mathrm{Dec}(\mathrm{Enc}(s_t))\big)$
      \State \textbf{Surprise:} obtain latents $(z_{t+1}, z'_{t+1})$ and set $\displaystyle \mathrm{Surprise}(s_t,a_t) \gets \mathrm{MSE}(z'_{t+1}, z_{t+1})$
      \State \textbf{NSS:} $\displaystyle \mathrm{NSS}(s_t,a_t) \gets \mathrm{Novelty}(s_t) + \mathrm{Surprise}(s_t,a_t)$
      \State \textbf{Intrinsic reward:} $r_{\text{int}} \gets \mathrm{NSS}(s_t,a_t)$
      \State \textbf{Total reward:} 
      $
      r_t \gets
      \begin{cases}
        r_{\text{ext}} + r_{\text{int}}, & \text{NSPER+R}\\
        r_{\text{ext}}, & \text{NSPER}
      \end{cases}
      $
      \State Push $(s_t,a_t,r_t,s_{t+1},\mathrm{NSS}(s_t,a_t))$ into $B$
      \State Set priority for new item: $\sigma_{\text{new}} \gets |\mathrm{NSS}(s_t,a_t)| + \varepsilon$
    
      \For{each update step}
        \State \textbf{Sample} indices $I$ of size $m$ with PER:
          \[
          p(i) = \frac{\sigma_i^{\alpha}}{\sum_{k} \sigma_k^{\alpha}}
          \]
        \State \textbf{IS weights:} $w(i) \gets \Big(\frac{1}{|B|\cdot p(i)}\Big)^{\beta}$ \; (optionally normalize by $\max_{j\in I} w(j)$)
        \State \textbf{Update critic/actor:} compute losses on batch, weight by $w(i)$, update $\theta,\Theta$
        \State \textbf{Refresh priorities:} for each $i\in I$, set $\sigma_i \gets |\mathrm{NSS}(s_i,a_i)| + \varepsilon$
      \EndFor
    \EndFor
    \end{algorithmic}
    \end{algorithm}

 \begin{figure*}[!t]
    \centering
    \begin{subfigure}[t]{0.19\textwidth}
        \includegraphics[width=\linewidth]{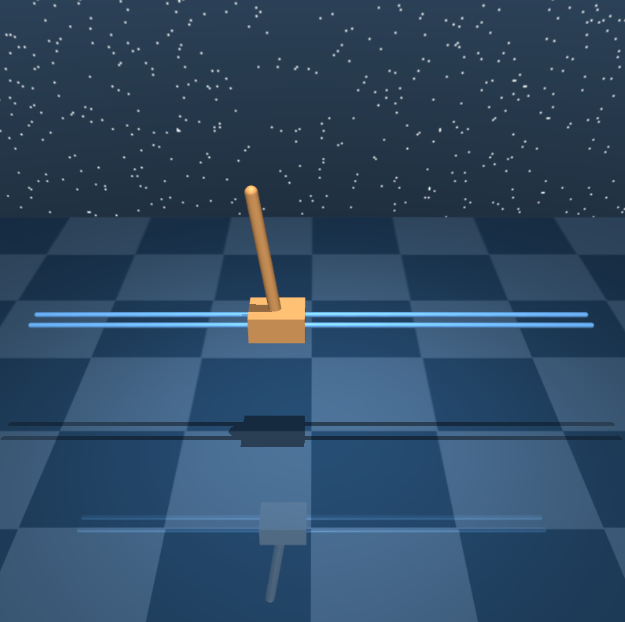}
        \caption{Cartpole-Balance}
        \label{fig:cartpole}
    \end{subfigure}%
    \hfill
    \begin{subfigure}[t]{0.19\textwidth}
        \includegraphics[width=\linewidth]{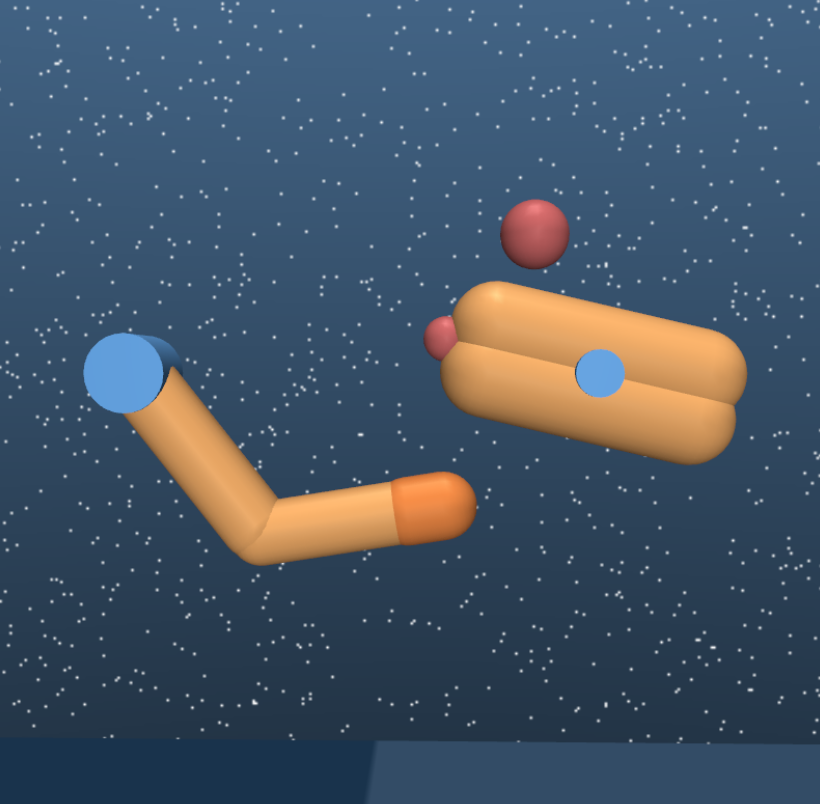}
        \caption{Finger-Spin}
        \label{fig:finger}
    \end{subfigure}%
    \hfill
    \begin{subfigure}[t]{0.19\textwidth}
        \includegraphics[width=\linewidth]{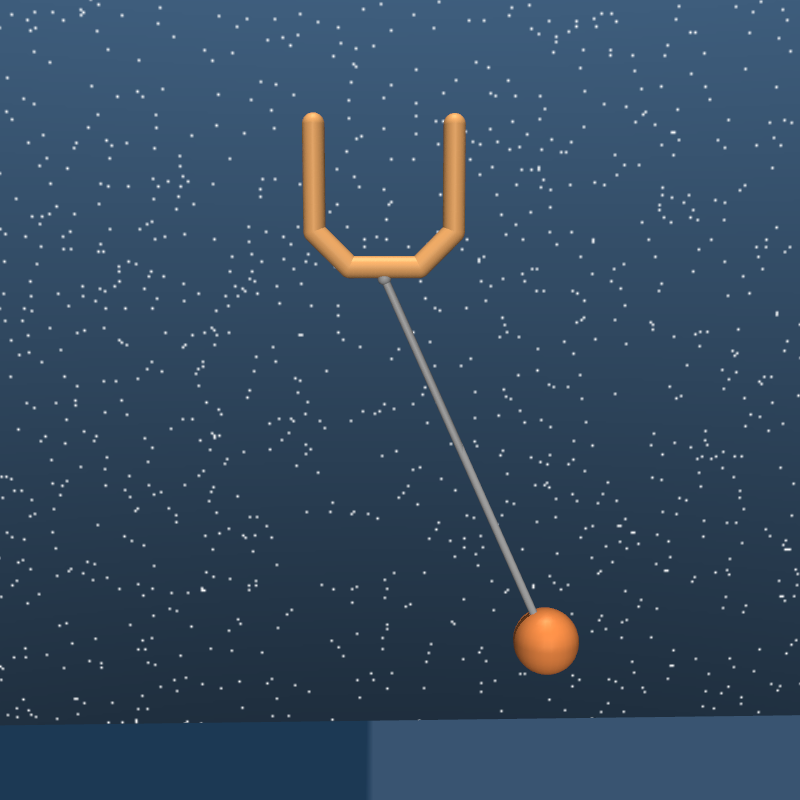}
        \caption{Ball-in-Cup}
        \label{fig:ball}
    \end{subfigure}%
    \hfill
    \begin{subfigure}[t]{0.19\textwidth}
        \includegraphics[width=\linewidth]{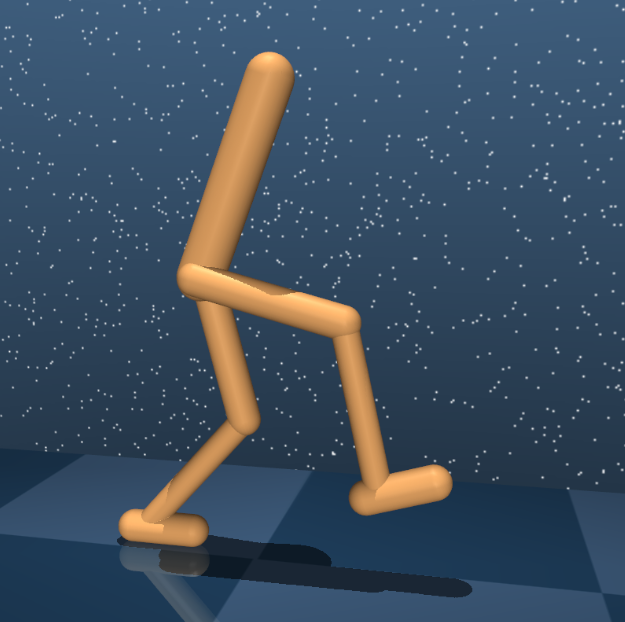}
        \caption{Walker-Walk}
        \label{fig:walker}
    \end{subfigure}%
    \hfill
    \begin{subfigure}[t]{0.19\textwidth}
        \includegraphics[width=\linewidth]{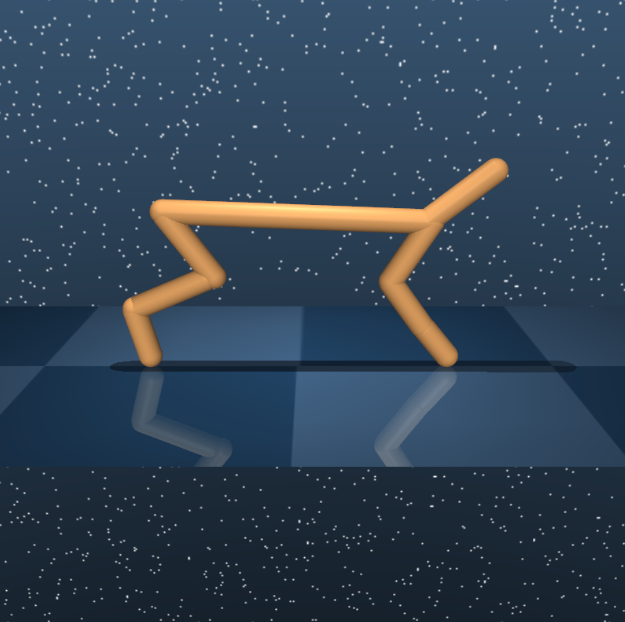}
        \caption{Cheetah-Run}
        \label{fig:cheetah}
    \end{subfigure}%
    \caption{Five continuous control tasks from the DeepMind Control Suite used in our evaluation.}
    \label{fig:dmcs_envs}
\end{figure*}  

\begin{table*}[htbp]
\centering
\caption{Summary of DeepMind Control Suite tasks used in our experiments.}
\label{tab:dmcs-tasks}
\resizebox{\textwidth}{!}{%
\begin{tabular}{l
                p{0.15\textwidth}
                p{0.26\textwidth}
                p{0.32\textwidth}
                p{0.12\textwidth}
                p{0.10\textwidth}}
\toprule
\textbf{Task} &
\textbf{Action Space} &
\textbf{Objective} &
\textbf{Primary Challenge} &
\textbf{Type} &
\textbf{Difficulty} \\
\midrule

Cartpole-Balance &
$[-1,1]^1$ &
Stabilize the pole by moving the cart horizontally. &
Underactuated and highly unstable dynamics where small errors rapidly lead to failure. &
Classic Control &
Low \\

Finger-Spin &
$[-1,1]^2$ &
Maintain continuous rotation of a free-floating object. &
Requires fine torque modulation to sustain spin without drift or loss of contact. &
Manipulation &
High \\

Ball-in-Cup &
$[-1,1]^2$ &
Swing the ball and capture it inside the cup. &
Oscillatory motion requires precise timing and well-coordinated control. &
Manipulation &
Medium--High \\

Walker-Walk &
$[-1,1]^6$ &
Achieve stable forward locomotion. &
Balancing and coordinating a bipedal body in an inherently unstable locomotion regime. &
Locomotion &
Medium \\

Cheetah-Run &
$[-1,1]^6$ &
Run forward at high speed. &
High velocity amplifies instability, requiring strong, consistent, and fast control. &
Locomotion &
Medium \\

\bottomrule
\end{tabular}}
\end{table*}
\begin{table*}[htbp]
\centering
\caption{Experimental setup for NSPER across DeepMind Control Suite environments.}
\label{tab:exp-setup}
\resizebox{\textwidth}{!}{%
\begin{tabular}{lp{0.75\textwidth}}
\toprule
\textbf{Parameter} & \textbf{Value (Description)} \\
\midrule
\multicolumn{2}{c}{\textbf{Environments}} \\
Observation & $84 \times 84 \times 3$ RGB images, with 3 consecutive frames stacked \\
Reward Scale & Maximum episodic return of 1000 per task \\
Evaluation Metric & Average cumulative reward per step (computed using extrinsic rewards) \\
\midrule
\multicolumn{2}{c}{\textbf{Implementation Details}} \\
Replay Buffer Capacity & $1{,}000{,}000$ transitions \\
Batch Size & 128 (Mini-batch sampled from replay buffer) \\
Actor Learning Rate & $1 \times 10^{-4}$ \\
Critic Learning Rate & $1 \times 10^{-3}$ \\
Prioritization Exponent $\alpha$ & 0.7 (Controls prioritization sharpness) \\
Importance-Sampling Exponent $\beta$ & 0.4 (Corrects sampling bias) \\
Training Steps & $1 \times 10^{6}$ (Total environment steps) \\
Evaluation Frequency & Every $1 \times 10^{4}$ steps \\
Evaluation Episodes & 10 (Episodes per evaluation checkpoint) \\
Random Seeds & 5 (Independent runs with different seeds for environment and initialization) \\
\bottomrule
\end{tabular}}
\end{table*}

\begin{table*}[!t]
\centering
\caption{Network architectures used in NSPER and NSPER+R.}
\label{tab:arch-setup}
\resizebox{\textwidth}{!}{%
\begin{tabular}{lcccc}
\toprule
\textbf{Component} & \textbf{Input} & \textbf{Layers / Units} & \textbf{Activations} & \textbf{Output} \\
\midrule
Encoder & $84\times84\times3$ image & 4 conv (32, $3\times3$) + FC(200) & ReLU (conv), $\tanh$ (FC) & Latent $z \in \mathbb{R}^{200}$ \\
Decoder & Latent $z$ & Deconv mirror of encoder & ReLU, Sigmoid (final) & Image reconstruction \\
Actor (TD3) & $z$ (opt. + vector obs) & FC(1024) → FC(1024) & ReLU, $\tanh$ (out) & Action $\in [-1,1]^d$ \\
Critics (TD3, $\times 2$) & $z \| a$ (opt. + vector obs) & FC(1024) → FC(1024) & ReLU & Scalar $Q$-value \\
Predictive Ensemble & $z_t \| a_t$ & FC(512) → FC(512) & ReLU & Pred. latent $\hat z_{t+1}$ \\
\bottomrule
\end{tabular}}
\end{table*}

\section{\textbf{EXPERIMENTS}}

    \begin{figure*}[htbp]
        \centering
        \begin{subfigure}[b]{0.32\textwidth}
            \includegraphics[width=\textwidth]{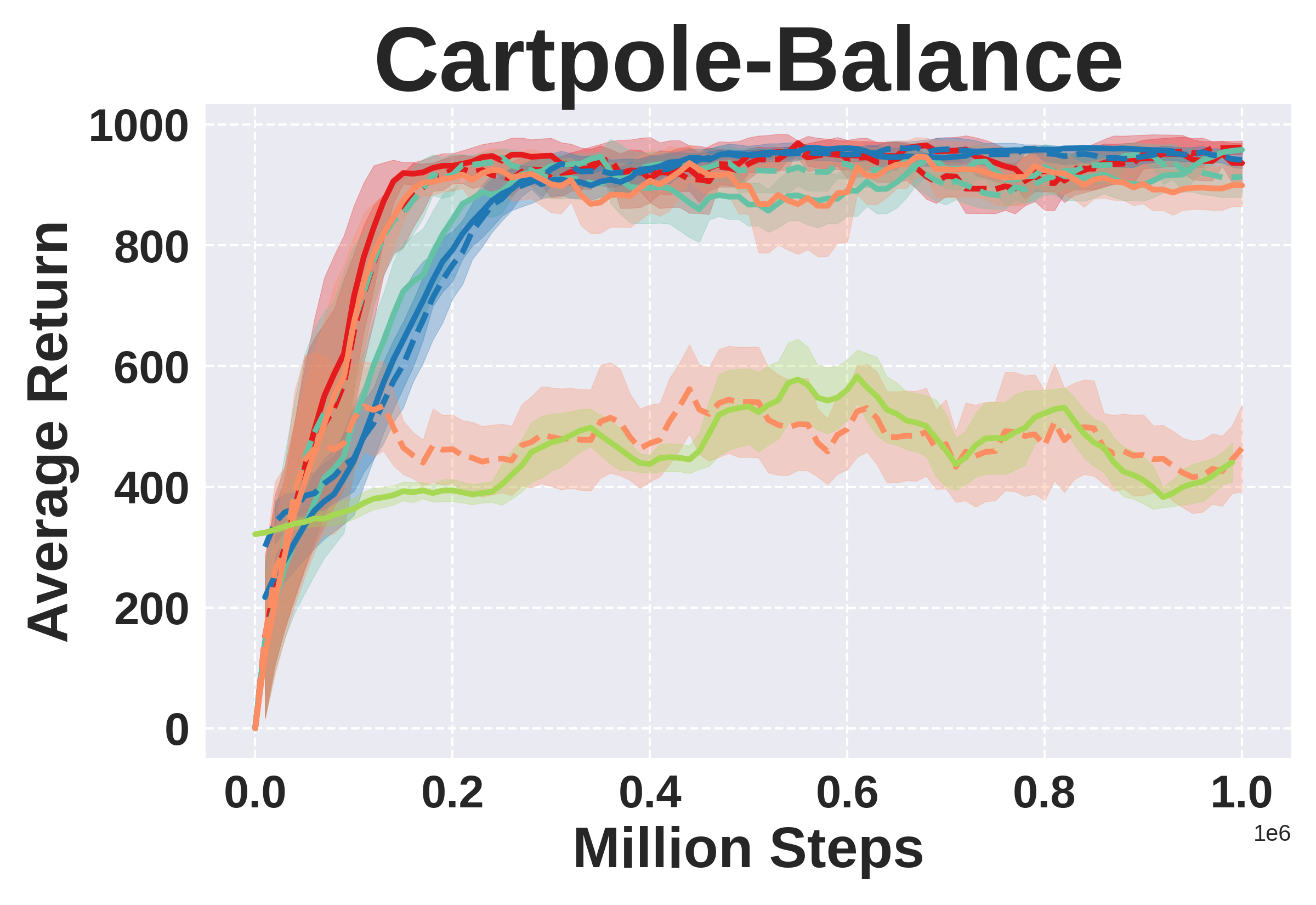}
            \caption{}
        \end{subfigure}
        \hfill
        \begin{subfigure}[b]{0.32\textwidth}
            \includegraphics[width=\textwidth]{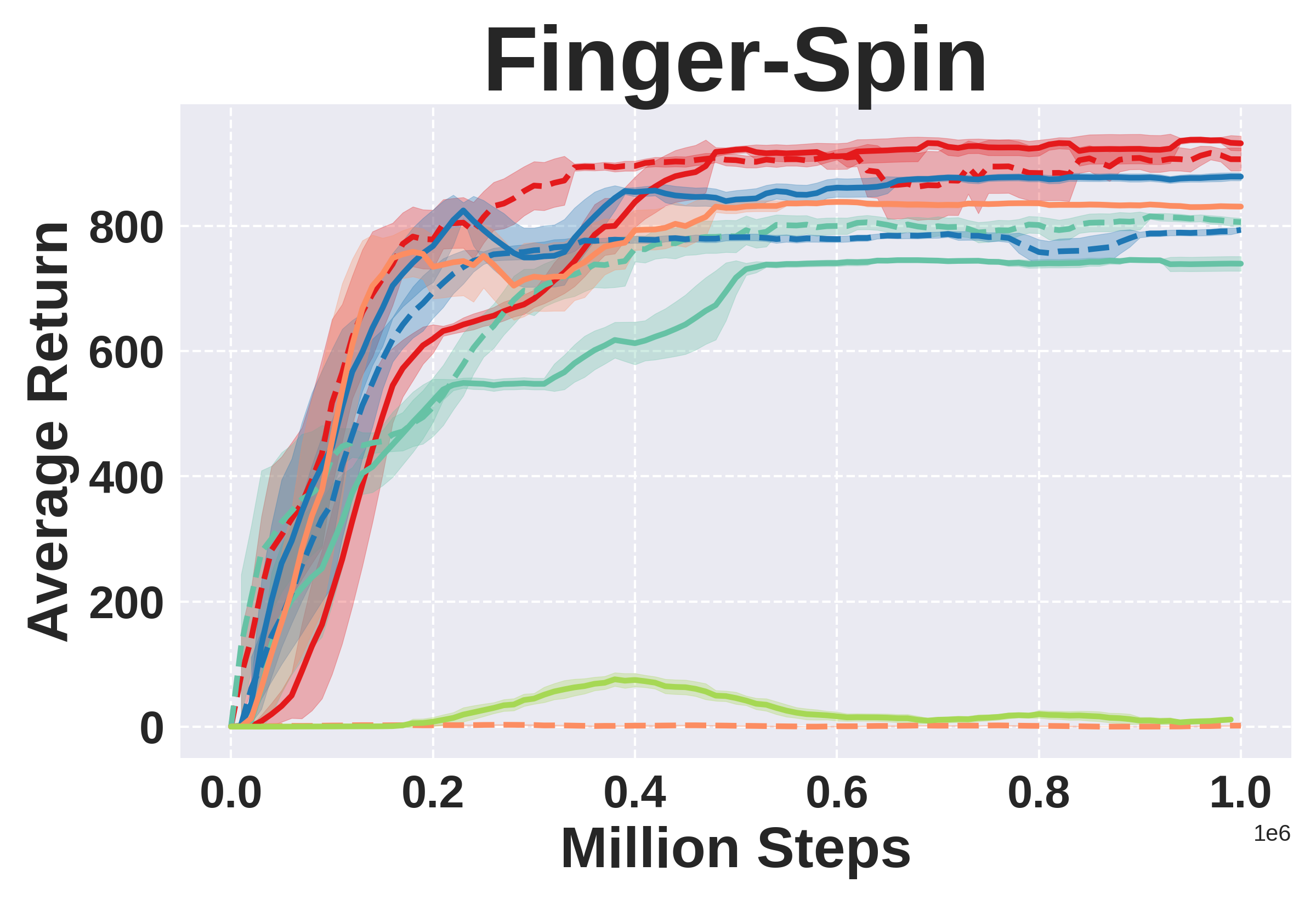}
            \caption{}
        \end{subfigure}
        \hfill
        \begin{subfigure}[b]{0.32\textwidth}
            \includegraphics[width=\textwidth]{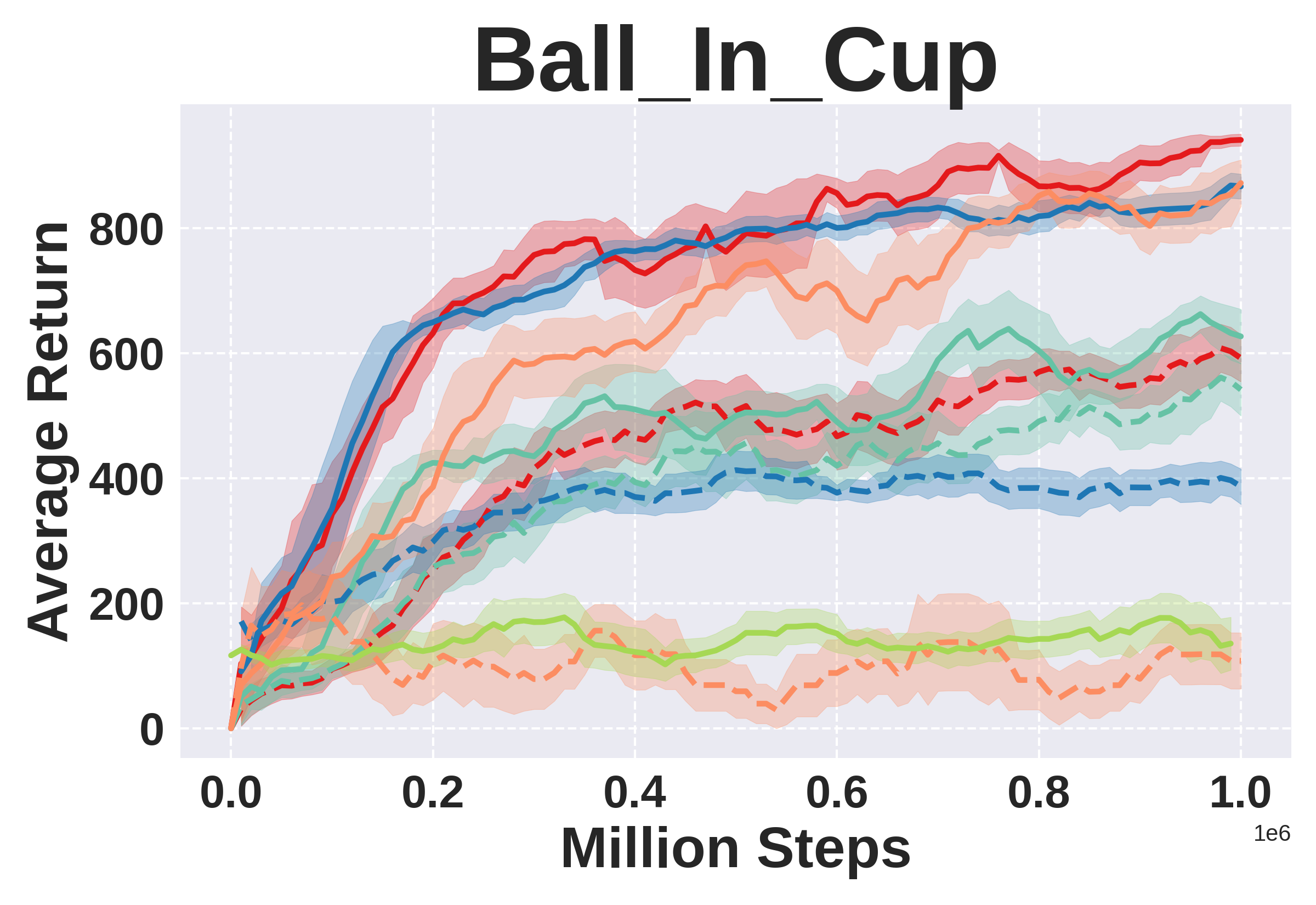}
            \caption{}
        \end{subfigure}
    
        \vspace{3pt} 
    
        \centering
        \begin{subfigure}[b]{0.32\textwidth}
            \centering
            \includegraphics[width=\textwidth]{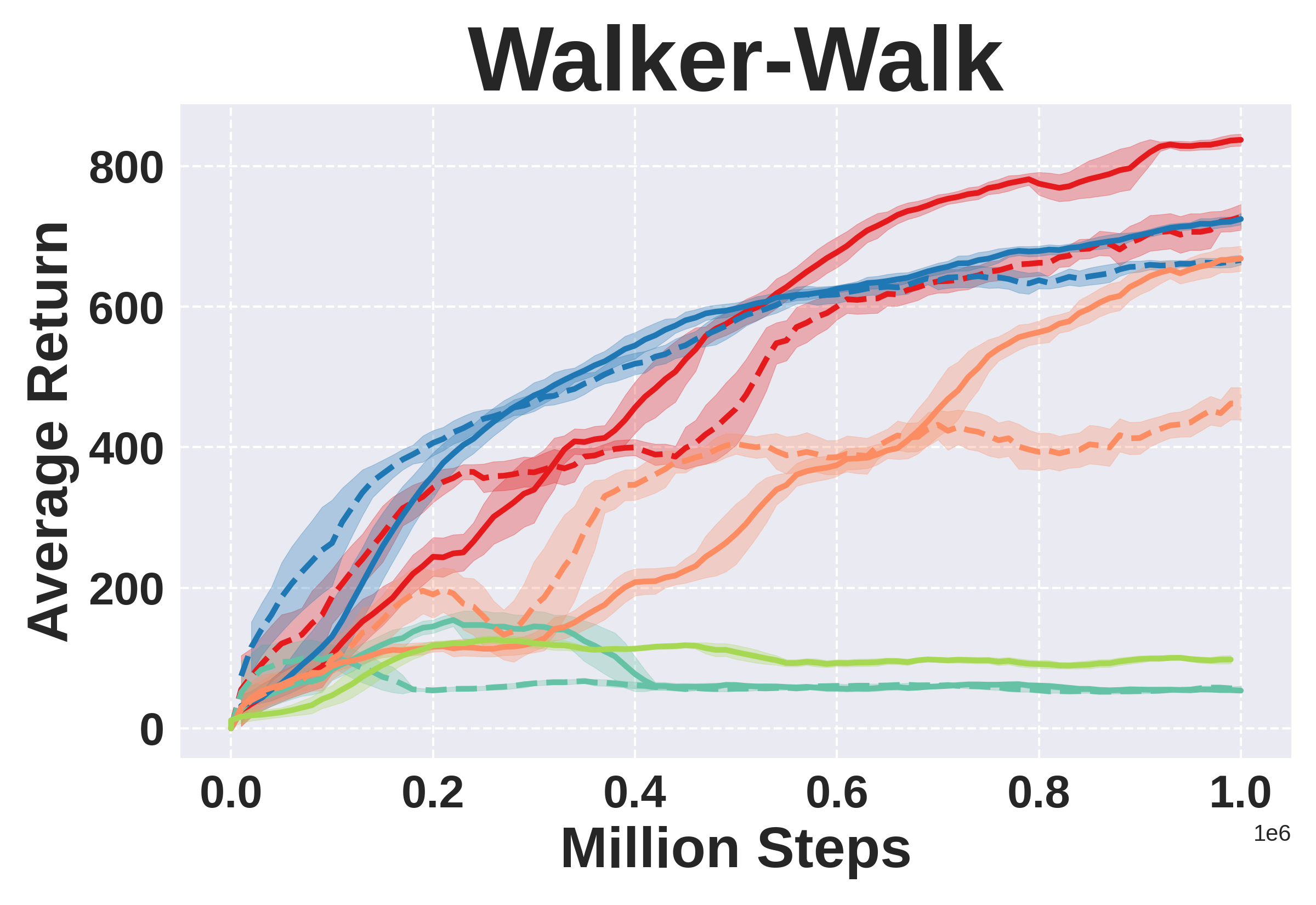}
            \caption{}
        \end{subfigure}
        \begin{subfigure}[b]{0.32\textwidth}
            \centering
            \includegraphics[width=\textwidth]{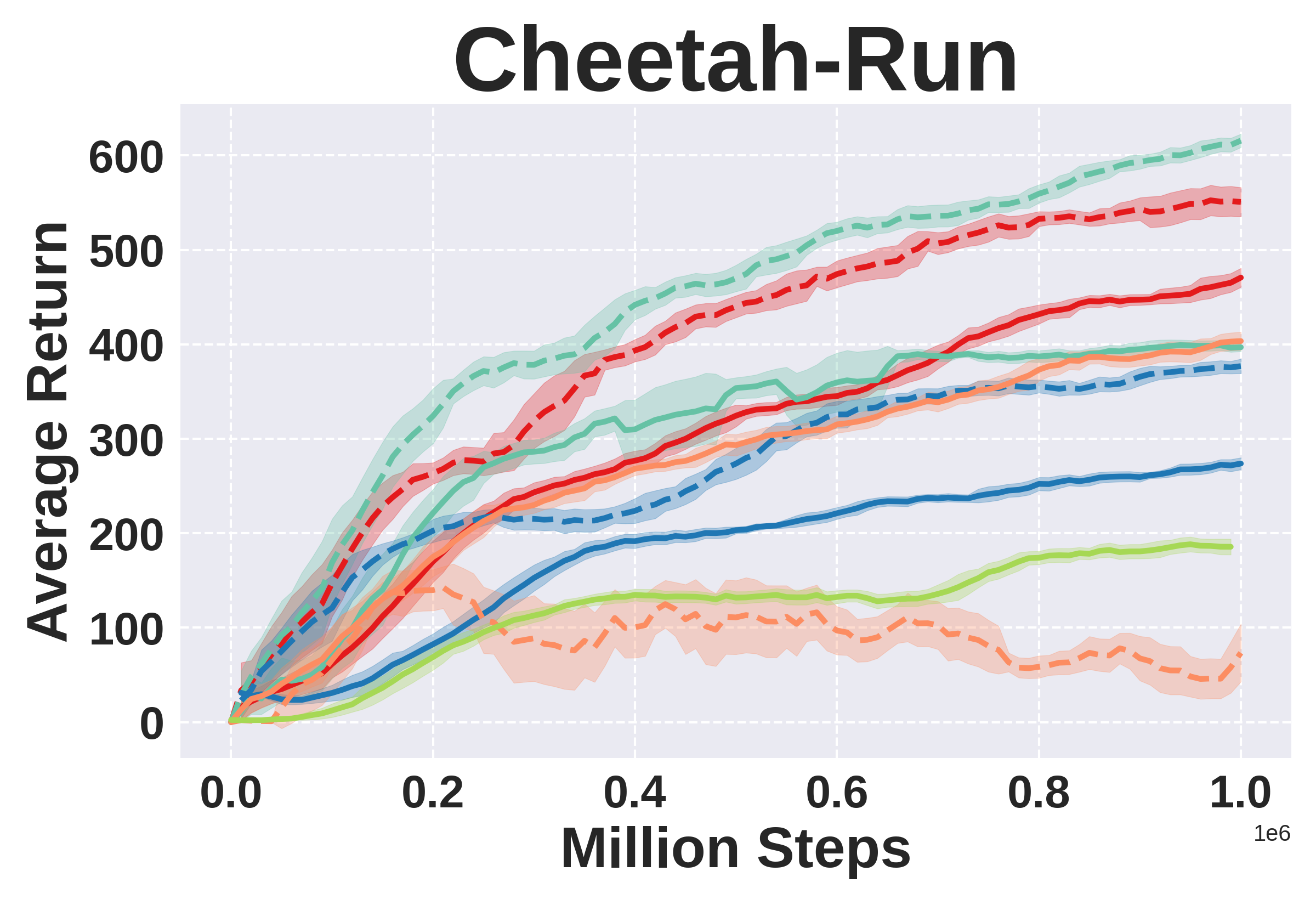}
            \caption{}
        \end{subfigure}
    
        \vspace{6pt} 
    
        \resizebox{\linewidth}{!}{%
\begin{tabular}{@{}l@{}}
  \textcolor{red}{\rule{0.025\linewidth}{0.15cm}} \scriptsize \textbf{NSPER+R} \hspace{1em}
  \textcolor{red}{\hdashrule[0.5ex][c]{0.03\linewidth}{0.1cm}{0.1cm 0.5mm}} \scriptsize \textbf{NSPER}  \hspace{1em}
  \textcolor{customgreen}{\rule{0.025\linewidth}{0.15cm}} \scriptsize TD-PER+R \hspace{1em}
  \textcolor{customgreen}{\hdashrule[0.5ex][c]{0.03\linewidth}{0.1cm}{0.1cm 0.5mm}} \scriptsize TD-PER \hspace{1em}
  \textcolor{customblue}{\rule{0.025\linewidth}{0.15cm}} \scriptsize RPE-PER+R \hspace{1em}
  \textcolor{customblue}{\hdashrule[0.5ex][c]{0.03\linewidth}{0.1cm}{0.1cm 0.5mm}} \scriptsize RPE-PER \hspace{1em}
  \textcolor{customorange}{\rule{0.025\linewidth}{0.15cm}} \scriptsize Uniform+R \hspace{1em}
  \textcolor{customorange}{\hdashrule[0.5ex][c]{0.03\linewidth}{0.1cm}{0.1cm 0.5mm}} \scriptsize Uniform \hspace{1em}
  \textcolor{CCLF}{\rule{0.025\linewidth}{0.15cm}} \scriptsize CCLF
\end{tabular}}

        \caption{Learning curves for continuous control tasks using PixelTD3. Solid lines represent the algorithm's performance with intrinsic rewards (+R), while dashed lines indicate performance without intrinsic rewards. Shaded regions denote the 95\% confidence interval across 5 runs.}
        \label{fig:Final-Results}
    \end{figure*}


This section describes the experimental setup, simulation tasks, algorithmic settings, and baseline methods used in our evaluation, and then presents an analysis of the learning performance of NSPER, NSPER+R, and competing approaches.

\subsection{\textbf{Experimental Setup}}

We evaluate our algorithms and baselines on five image-based continuous control tasks from the DeepMind Control Suite (DMC)~\cite{tassa2018deepmind}: \textbf{Cartpole-Balance}, \textbf{Finger-Spin}, \textbf{Ball-in-Cup}, \textbf{Walker-Walk}, and \textbf{Cheetah-Run}. These environments constitute the standard benchmark suite for visual reinforcement learning and have been widely used in prior pixel-based studies~\cite{kostrikov2020image,laskin2020curl,valencia2024image}. This set provides representative coverage of balance, manipulation, and locomotion tasks and offers sufficient diversity for assessing prioritization strategies in image-based continuous control. Visual examples of the environments are given in Fig.~\ref{fig:dmcs_envs}, and their key properties are summarised in Table~\ref{tab:dmcs-tasks}.

All algorithms are trained under a unified experimental configuration following established practice in continuous-control evaluation. Training settings, optimization parameters, evaluation schedules, and replay specifications are listed in Table~\ref{tab:exp-setup}. Using identical configurations across all tasks ensures that performance differences arise from the prioritization mechanism rather than from hyperparameter choices or implementation bias. NSPER and NSPER+R employ a shared latent-space architecture for representation learning, prediction, and action-value estimation. The components of this architecture are outlined in Table~\ref{tab:arch-setup}.

 \subsection{\textbf{Experimental Approach}}

   To evaluate the effectiveness of our approach, we consider two proposed variants, NSPER and NSPER+R:  

    \textbf{NSPER:} Uses novelty and surprise solely for replay prioritization, isolating their effect on experience selection.  
    
    \textbf{NSPER+R:} Extends NSPER by also incorporating novelty and surprise as intrinsic rewards, combined with extrinsic rewards~\eqref{eq:11}, allowing us to assess their joint impact on prioritization and exploration.  
    
    \textbf{Baseline Algorithms:} For comparison, we evaluate a range of established methods within the PixelTD3 framework, which differ in their replay strategies and use of intrinsic rewards. All implementations share the same codebase for fairness. Although CCLF was originally proposed with a pixel-based SAC backbone, we adapt it to PixelTD3 for consistency. The baselines include:

    \begin{enumerate}
    \item \textbf{Uniform}: Uniformly samples transitions from the buffer without PER or intrinsic rewards \cite{fujimoto2018addressing, fedus2020revisiting}.
    \item \textbf{Uniform+R (NaSATD3)}: Uniform sampling with intrinsic rewards, equivalent to PixelTD3+R \cite{valencia2024image}.
    \item \textbf{TD-PER}: PER based on TD error \cite{schaul2015prioritized}, without intrinsic rewards.
    \item \textbf{TD-PER+R}: TD-PER augmented with intrinsic rewards.
    \item \textbf{RPE-PER}: PER based on reward prediction error (RPE) \cite{yamani2025reward}, without intrinsic rewards.
    \item \textbf{RPE-PER+R}: RPE-PER extended with intrinsic rewards.     
    \item \textbf{CCLF}: A contrastive learning-based prioritization method with intrinsic rewards to improve exploration and representation learning \cite{sun2022cclf}.
\end{enumerate}

This setup enables a comprehensive evaluation of NSPER and NSPER+R against diverse baselines that vary in their use of prioritization strategies and intrinsic rewards, providing a robust comparison in image-based RL.

    \subsection{\textbf{Learning Efficiency Analysis}}

        The learning curves in Fig. \ref{fig:Final-Results} illustrate the learning efficiency of NSPER and baseline methods across five challenging DeepMind continuous control environments. These curves depict the average evaluation reward over training steps, providing insights into convergence speed and performance differences among NSPER, NSPER+R, and the baseline algorithms.  
    
        Dashed-line plots represent PixelTD3 with different replay buffer sampling methods but without intrinsic rewards, while solid-line plots of the same color correspond to the same sampling methods with intrinsic rewards. These separate plots directly compare how sampling strategies influence performance with and without intrinsic rewards. 

        It is crucial to note that Baseline methods did not always replicate the performance reported in prior work. Such deviations are not uncommon in deep reinforcement learning and can arise from factors such as environmental stochasticity, initialization randomness, and subtle implementation or runtime differences. As emphasized by Henderson et al. \cite{henderson2018deep}, ensuring fair comparison is essential; in our case, the relative ranking of methods remained consistent across multiple runs, supporting the robustness of our evaluations.

  
 \section{\textbf{Results and discussion}}
       Performance results across various tasks in  Fig. \ref{fig:Final-Results} show that NSPER (red plot) results are better than baseline algorithms. It enhances PixelTD3's performance on four of five tasks while achieving comparable results on the fifth task to the best-performing methods among image-based RL approaches. Our evaluation highlights the crucial role of novelty and surprise signals in improving learning efficiency. In contrast, Uniform sampling (orange plot) exhibits suboptimal performance across all tasks, demonstrating that relying solely on a uniform experience buffer is insufficient. These findings suggest that PixelTD3 could significantly benefit from integrating prioritized sampling or intrinsic reward mechanisms, even in relatively simple environments.

       Incorporating prioritization strategies alongside intrinsic rewards improves performance across nearly all algorithms, highlighting the advantages of these techniques in fostering more robust and adaptive learning. In particular, NSPER+R achieves the best results among the methods. While CCLF (light green plot) consistently outperforms Uniform across all tasks, its overall performance remains unremarkable. This suggests that CCLF's prioritization strategy differs fundamentally from that of a prioritization buffer, and, as the results indicate, it does not integrate well with PixelTD3.
        
       In addition, the comparable performance of the RPE-PER algorithm (blue plot) to NSPER across all tasks demonstrates its potential as an effective prioritization method. However, NSPER consistently achieves better results. These findings indicate that using a complex prioritization scale in image-based RL can be beneficial, while also reaffirming that novelty and surprise signals are stronger heuristics for selecting informative experiences, ultimately contributing to a more sample-efficient learning process.

         \begin{figure*}[htbp]
        \centering
        \begin{subfigure}[b]{0.32\textwidth}
            \includegraphics[width=\textwidth]{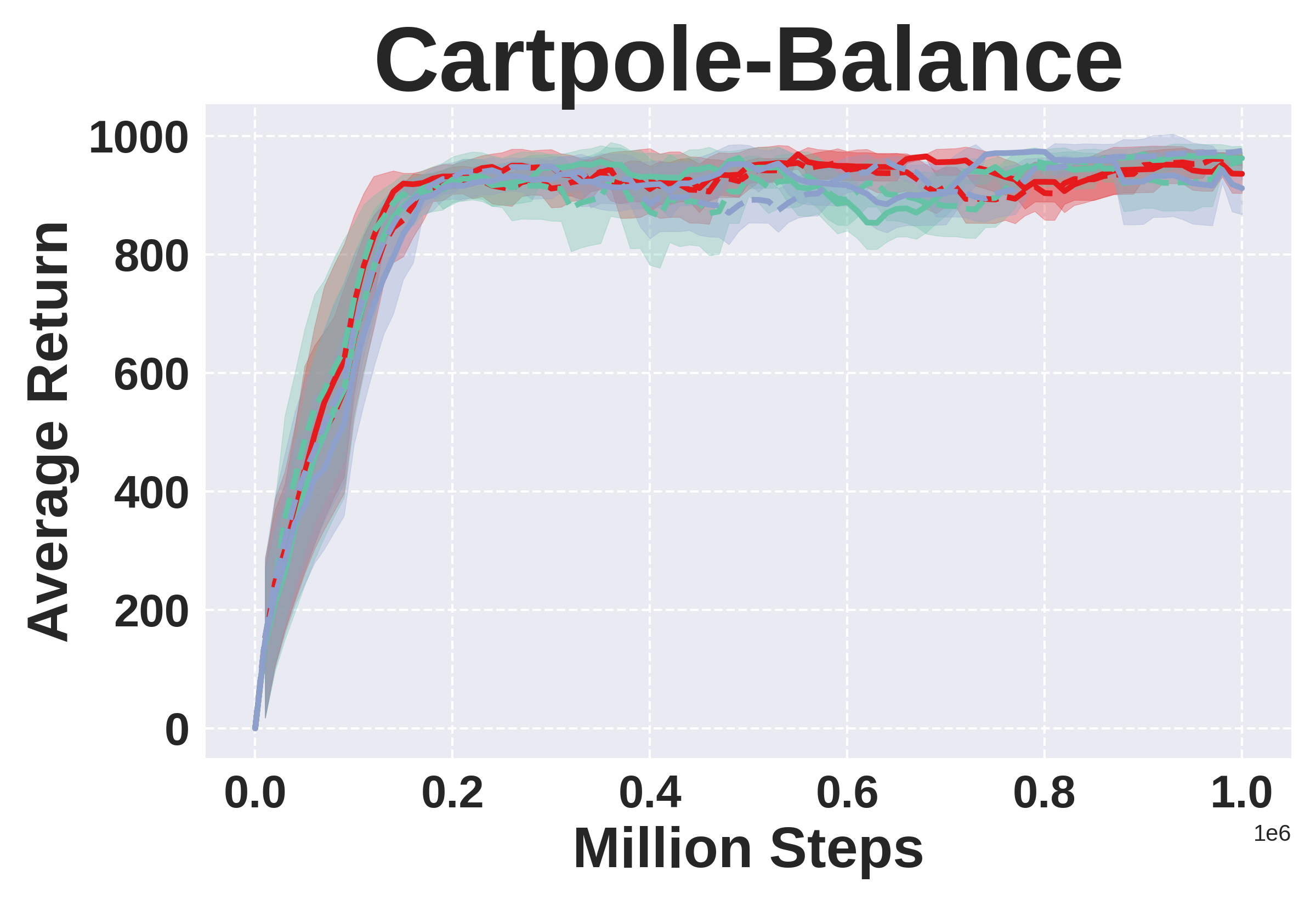}
            \caption{}
        \end{subfigure}
        \hfill
        \begin{subfigure}[b]{0.32\textwidth}
            \includegraphics[width=\textwidth]{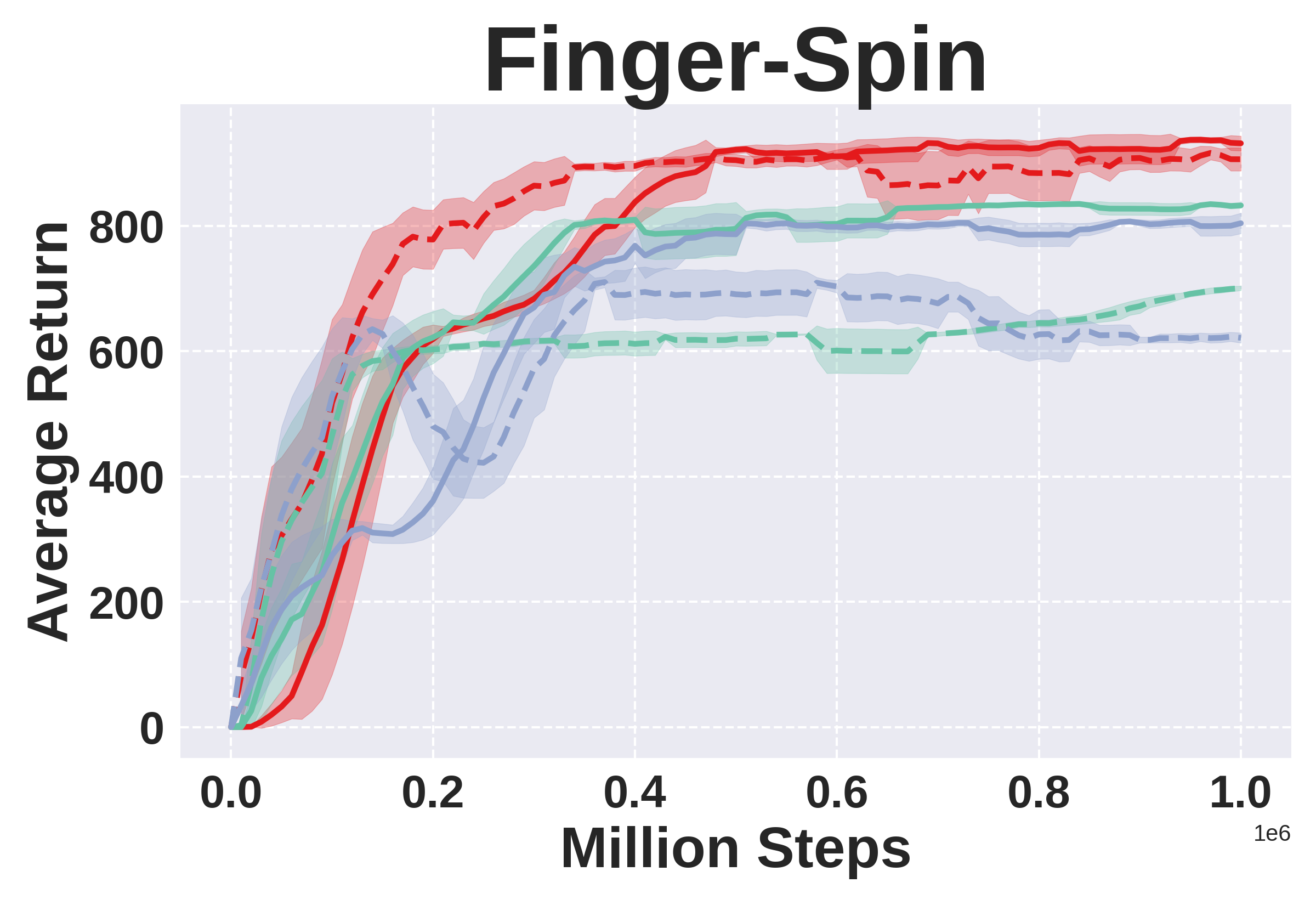}
            \caption{}
        \end{subfigure}
        \hfill
        \begin{subfigure}[b]{0.32\textwidth}
            \includegraphics[width=\textwidth]{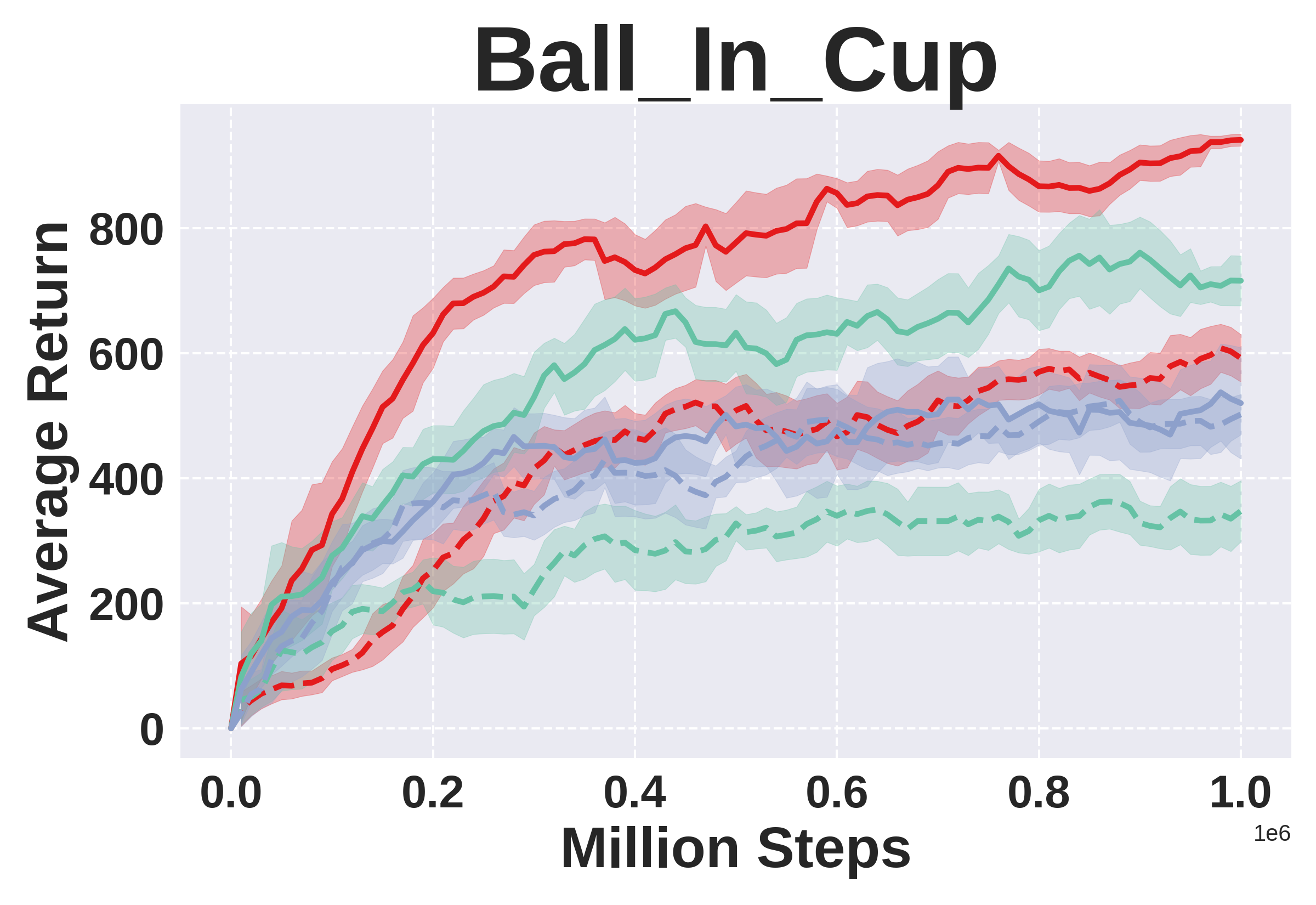}
            \caption{}
        \end{subfigure}
    
    
        \centering
        \begin{subfigure}[b]{0.32\textwidth}
            \centering
            \includegraphics[width=\textwidth]{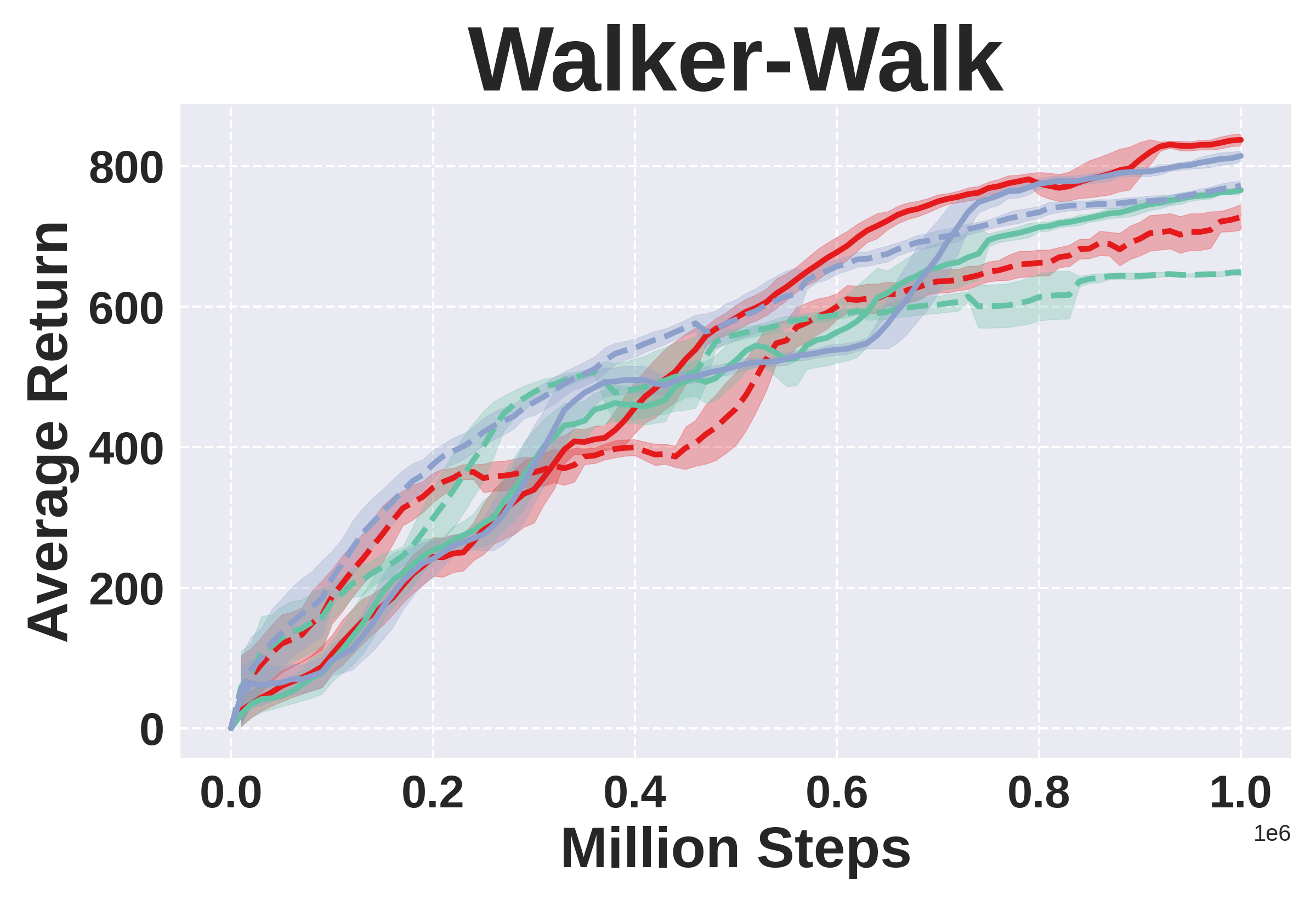}
            \caption{}
        \end{subfigure}
        \begin{subfigure}[b]{0.32\textwidth}
            \centering
            \includegraphics[width=\textwidth]{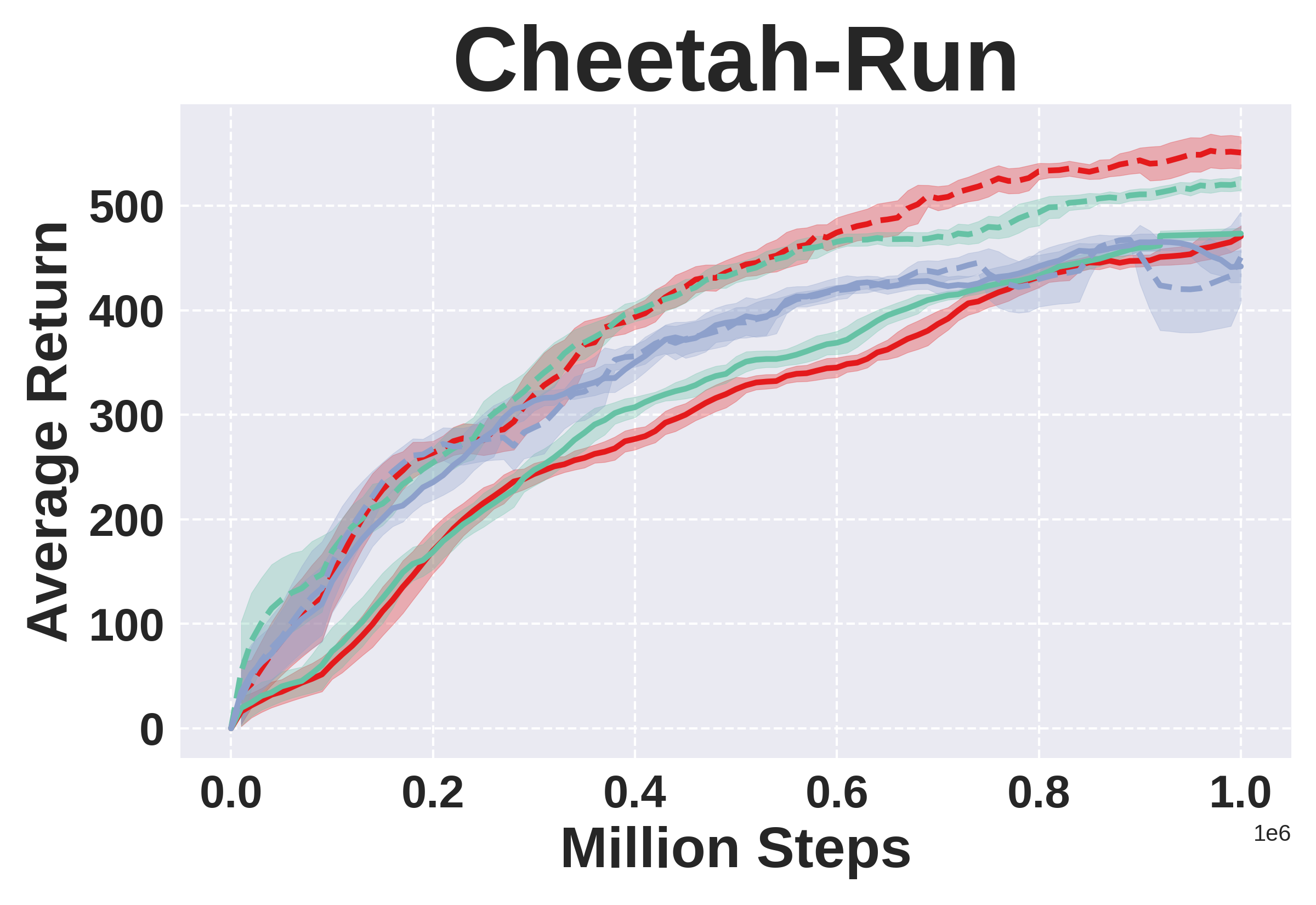}
            \caption{}
        \end{subfigure}
    
    
        \begin{tabular}{@{}l@{}}
            \textcolor{red}{\rule{0.025\linewidth}{0.15cm}} \scriptsize \textbf{NSPER+R} 
            \hspace{1em}
           \textcolor{red}{\hdashrule[0.5ex][c]{0.03\linewidth}{0.1cm}{0.1cm 0.5mm}} \scriptsize \textbf{NSPER}  
            \hspace{1em}  
            
            \textcolor{customgreen}{\rule{0.025\linewidth}{0.15cm}} \scriptsize \textbf{NoveltyPER+R} 
            \hspace{1em}
            \textcolor{customgreen}{\hdashrule[0.5ex][c]{0.03\linewidth}{0.1cm}{0.1cm 0.5mm}} \scriptsize \textbf{NoveltyPER}
            \hspace{1em}
            \textcolor{customgrey}{\rule{0.025\linewidth}{0.15cm}} \scriptsize \textbf{SurprisePER+R} 
            \hspace{1em}
            \textcolor{customgrey}{\hdashrule[0.5ex][c]{0.03\linewidth}{0.1cm}{0.1cm 0.5mm}} \scriptsize \textbf{SurprisePER} 
        \end{tabular}
        \caption{Ablation results comparing experience prioritization based on novelty, surprise, and their combination, with and without intrinsic rewards. Shaded regions represent the 95\% confidence interval over five runs.}
        
        \label{fig:Ablition-Results}
    \end{figure*}
        
    \subsection{\textbf{Task-Specific Analysis}}
        
        This section examines the performance of different methods across individual tasks, highlighting their strengths and limitations in each environment.
        
        In Cartpole-Balance (Fig. \ref{fig:Final-Results}, plot a), a classic control task requiring a cart to balance a pole, all methods using either a prioritized buffer or intrinsic rewards perform well. This suggests that PixelTD3 benefits from PER or intrinsic rewards, even in simpler environments.
        
        The Finger-Spin task, which involves manipulating an object with a simulated finger, presents challenges due to complex contact dynamics and sparse rewards. As shown in Fig.~\ref{fig:Final-Results}, plot b, NSPER variants outperform other methods, emphasizing the importance of novelty and surprise signals for learning efficiency. NSPER+R, which integrates intrinsic rewards with PER, further enhances performance. Comparing TD-PER (prioritization based on TD error) with Uniform+R (intrinsic rewards without prioritization) reveals that intrinsic rewards yield better results in this environment than PER alone. However, TD-PER+R underperforms relative to TD-PER, suggesting that conflicting signals from the TD error and novelty-surprise signals can hinder efficiency. In contrast, NSPER+R aligns both prioritization and intrinsic rewards using the same features, thereby improving sample efficiency.
        
        The Ball-in-Cup task, characterized by sparse rewards granted only upon successfully catching the ball, provides another challenging test. In (Fig. \ref{fig:Final-Results}, plot c), NSPER+R outperforms other approaches, demonstrating the benefits of combining novelty and surprise for prioritization with intrinsic rewards. While Uniform+R performs well, NSPER+R and RPE-PER+R achieve consistently better results, reinforcing that properly aligning prioritization with intrinsic rewards enhances performance.
        
        The Walker-Walk environment, which requires controlling a bipedal walker to maintain dynamic balance, presents a complex, multi-stage control challenge despite offering dense rewards. (Fig. \ref{fig:Final-Results}, plot d) shows that NSPER+R and RPE-PER+R improve performance, highlighting the importance of context-aware experience selection. In contrast, TD error-based PER leads to suboptimal results even when combined with intrinsic rewards. This suggests that prioritizing solely based on high TD error can misguide learning, as such experiences may not always be the most informative.
        
        In Cheetah-Run (Fig. \ref{fig:Final-Results}, plot e), TD-PER performs best, but NSPER closely follows, demonstrating its adaptability to high-speed locomotion tasks. This result indicates that prioritization strategies should be tailored to specific environments, as more straightforward prioritization methods can sometimes be more effective. However, TD-PER exhibits inconsistency, performing well in Cheetah-Run but poorly in Walker-Walk, as noted in prior studies \cite{pan2022understanding}. In contrast, NSPER maintains stable performance across diverse tasks, suggesting it provides a more reliable prioritization strategy.
        
\subsection{\textbf{Ablation Study: The Role of Novelty and Surprise in Prioritization}}

    This study evaluates various prioritization strategies to isolate the individual and combined effects of novelty and surprise. NoveltyPER prioritizes experiences based solely on novelty, SurprisePER based solely on surprise, and NSPER jointly leverages both signals. Solid curves denote variants that additionally use intrinsic rewards, while dashed curves show performance without them.
    As shown in Fig.~\ref{fig:Ablition-Results}, prioritization based on both novelty and surprise outperforms using either signal alone. This combination improves experience selection and, when also incorporated as intrinsic rewards, encourages more effective exploration. The results highlight the complementary nature of novelty and surprise in RL. Furthermore, integrating these signals as intrinsic rewards yields additional performance gains by promoting a more balanced selection of experiences, ultimately enhancing policy learning.
    
    In the Cheetah-Run task, prioritization alone is sufficient, and incorporating surprise as an intrinsic reward negatively impacts performance. Another key observation is that NoveltyPER outperforms SurprisePER in most tasks, suggesting that novelty-based prioritization, which encourages exploration of diverse states, is more beneficial than surprise-based prioritization, which focuses on unexpected transitions. Additionally, the performance of SurprisePER and SurprisePER+R remains similar in most cases, indicating that adding surprise as an intrinsic reward does not significantly contribute to performance improvements. However, when both novelty and surprise are used in prioritization and intrinsic rewards, the agent achieves the best overall performance, demonstrating their complementary roles in improving learning efficiency.
\section{\textbf{Conclusion}}

   In this work, we addressed the challenge of sample inefficiency in image-based RL by introducing NSPER and its extension NSPER+R, which integrate novelty and surprise as signals for replay prioritization and, in the case of NSPER+R, also as intrinsic rewards. This dual mechanism improves both replay quality and exploration, leading to faster convergence and more generalizable policies. Through an extensive evaluation of the DeepMind Control Suite using the PixelTD3 backbone, we showed that NSPER and NSPER+R  outperform existing baselines, improving training efficiency in complex image-based tasks. Ablation studies further confirmed the complementary contributions of novelty and surprise to prioritization and exploration. Our research highlights the importance of using internal stimuli and enhanced sampling buffers to improve RL systems.

   Future work should explore additional intrinsic motivators and adaptively weight novelty and surprise. Our ablation study indicates that these signals impact learning efficiency differently, suggesting that dynamically adjusting their balance could refine the agent's exploration strategy.

\bibliographystyle{IEEEtran}
\bibliography{bibliography}
\end{document}